\pdfoutput=1
\PassOptionsToPackage{table,xcdraw}{xcolor}
\documentclass[11pt]{article}

\usepackage[preprint]{acl}

\usepackage{times}
\usepackage{latexsym}

\usepackage[T1]{fontenc}
\usepackage[utf8]{inputenc}

\usepackage{microtype}

\usepackage{inconsolata}

\usepackage{graphicx}
\usepackage{algorithm}
\usepackage{algpseudocode}
\usepackage{amsmath}
\usepackage{booktabs}
\usepackage{multirow}
\usepackage{subcaption}
\newcommand{\methode}{\textsc{LiTe}}

\title{\methode: \textbf{L}LM-\textbf{I}mpelled efficient \textbf{T}axonomy \textbf{E}valuation}

\author{
 \textbf{Lin Zhang\textsuperscript{1}},
 \textbf{Zhouhong Gu\textsuperscript{1}},
 \textbf{Suhang Zheng\textsuperscript{2}},
 \textbf{Tao Wang\textsuperscript{2}},
 \textbf{Tianyu Li\textsuperscript{2}},
 \\
 \textbf{Hongwei Feng\textsuperscript{1,}\thanks{Corresponding authors.}},
 \textbf{Yanghua Xiao\textsuperscript{1,}\footnotemark[1]},
\\
\\
 \textsuperscript{1}Shanghai Key Laboratory of Data Science, School of Computer Science, Fudan University,
 \\
 \textsuperscript{2}Alibaba Group
\\
\texttt{\{linzhang22, zhgu22\}@m.fudan.edu.cn,}\\
\texttt{\{hwfeng, shawyh\}@fudan.edu.cn,}\\
\texttt{\{suhang.zhengsh, shayue.wt, qianchuan.lty\}@alibaba-inc.com}
}

\begin{document}
\maketitle

\begin{abstract}

This paper presents \methode, an LLM-based evaluation method designed for efficient and flexible assessment of taxonomy quality. To address challenges in large-scale taxonomy evaluation, such as efficiency, fairness, and consistency, \methode~adopts a top-down hierarchical evaluation strategy, breaking down the taxonomy into manageable substructures and ensuring result reliability through cross-validation and standardized input formats. \methode~also introduces a penalty mechanism to handle extreme cases and provides both quantitative performance analysis and qualitative insights by integrating evaluation metrics closely aligned with task objectives. Experimental results show that \methode~demonstrates high reliability in complex evaluation tasks, effectively identifying semantic errors, logical contradictions, and structural flaws in taxonomies, while offering directions for improvement.
Code is available at \url{https://github.com/Zhang-l-i-n/TAXONOMY_DETECT}.
\end{abstract}

\section{Introduction}
Taxonomy plays a crucial role in knowledge organization, information retrieval, and task understanding, primarily through establishing hierarchical relationships between concepts \cite{Welty2001Supporting, 10.1145/3289600.3290972}. 
However, diverse application domains and scenarios impose varying requirements on taxonomic structure and content, presenting significant challenges in taxonomy construction and evaluation \cite{Glass1995Contemporary,Bordea2020Evaluation}. 
Although taxonomy evaluation is essential for quality assurance and continuous optimization, existing evaluation methodologies exhibit notable limitations in meeting practical application requirements.

Traditional taxonomy evaluation approaches can be broadly categorized into two paradigms: 
one focuses on the validity of local hypernym-hyponym relationships \cite{Jiang2022TaxoEnrich:, shen2020taxoexpan,gu2023gantee,cheng2022learning}, while the other examines the overall taxonomic structure \cite{lozano2004ontometric,acosta2018detecting,unterkalmsteiner2023compendium,amith2018assessing,yu2015learning,gao2023automatic,watts1998collective}.
The former enables precise assessment of specific conceptual relationships but neglects the holistic and systematic nature of taxonomies, whereas the latter captures macroscopic structural characteristics but fails to reflect the quality of constituent conceptual relationships.
This binary evaluation paradigm inadequately addresses the complexity of taxonomies.

Effective taxonomy evaluation should encompass both microscopic and macroscopic perspectives, as taxonomies comprise not only numerous concepts but also multidimensional relationship networks, including vertical hierarchical relationships and horizontal semantic associations \cite{Jiang2022TaxoEnrich:}.
Evaluating such complex structures requires simultaneous consideration of multiple dimensions: structural consistency, semantic coherence, and hierarchical validity.
Furthermore, the challenge of precisely aligning taxonomic structures with specific application requirements becomes increasingly critical, given the diverse demands across different scenarios \cite{Liu2024An}.

Recent advances in Large Language Models (LLMs) offer novel possibilities for addressing these complex evaluation challenges. Through training on extensive corpora, LLMs have demonstrated remarkable capabilities in understanding complex structures \cite{gu2024structext}, semantics \cite{gu2024xiezhi,huang2023c}, and contextual information \cite{openai2023gpt4}. These characteristics position LLMs as potential solutions for simultaneously addressing both microscopic and macroscopic aspects of taxonomy evaluation while adaptively accommodating diverse application requirements \cite{Raiaan2024A}.

Nevertheless, direct application of LLMs for taxonomy evaluation presents several challenges. First, LLMs' output sensitivity to minor input variations may lead to evaluation instability, particularly in iterative taxonomy assessment scenarios. Second, computational efficiency becomes particularly problematic in large-scale taxonomy evaluation, where reassessing all nodes after each iteration incurs substantial computational overhead \cite{Upadhyay2024LLMs}.

Given the potential and challenges of LLMs in taxonomy evaluation, developing an evaluation framework that leverages LLMs' advantages while addressing their limitations becomes paramount. Such a framework must ensure comprehensive and accurate assessment while considering evaluation efficiency and consistency. Through well-designed evaluation strategies and optimization techniques, it becomes feasible to achieve efficient, fair, and comprehensive evaluation of large-scale, complex taxonomies, thereby advancing taxonomic applications across various domains.

To address these challenges, we propose \textbf{L}LM-\textbf{I}mpelled \textbf{T}axonomy \textbf{E}valuation for Efficiency(\textbf{\methode}). 
Our approach incorporates a top-down hierarchical evaluation strategy, partitioning Taxonomies into manageable substructures to enhance evaluation efficiency. 
Additionally, we introduce penalty mechanisms to mitigate extreme scoring variations, ensuring fairness and reliability across different Taxonomy levels. 
Most importantly, \methode~integrates evaluation metrics that align closely with task-oriented objectives, enabling both quantitative performance analysis and qualitative insights into semantic and structural coherence. 
Through these approaches, \methode~systematically identifies weaknesses in Taxonomies and provides actionable recommendations for optimization and iteration.

By introducing this LLM-driven evaluation framework, we improve the accuracy and practical use of taxonomy quality evaluation.
Our experimental results demonstrate the defects of taxonomies under different construction methods (shown in Sec.\ref{ana:taxo}), showcasing the effectiveness of \methode~in advancing Taxonomy evaluation and refinement.

\section{Related work}
% 现有的分类法评价方法主要集中在基于结构和聚类的指标。这些方法通常测量Taxonomy的结构和预定义的基础真值标签之间的一致性，强调统计一致性和结构完整性。然而，这些方法很难解释特定于领域的业务语言的复杂性和语义的细微差别，这对于确保Taxonomy在实际场景中的实际适用性至关重要。此外，传统的聚类方法通常缺乏对高维或非结构化数据的适应性，限制了它们在需要对上下文语义和业务目标有细微理解的动态现实环境中的有效性。

Existing methods for evaluating Taxonomy primarily focus on structural and clustering-based metrics~\cite{Ng2017Classifying,Ciccarino2024A}, such as Normalized Mutual Information (NMI)~\cite{Kvålseth2017On} and Adjusted Rand Index (ARI)~\cite{D’Ambrosio2020Adjusted,Santos2009On}. 
These approaches typically measure the alignment between a Taxonomy’s structure and predefined ground truth labels, emphasizing statistical consistency and structural integrity. 
However, such methods struggle to account for the complexity of domain-specific business language and semantic nuances, which are crucial for ensuring Taxonomy’s practical applicability in real-world scenarios. 
Furthermore, traditional clustering methods often lack adaptability to high-dimensional or unstructured data, limiting their effectiveness in dynamic, real-world contexts that require a nuanced understanding of contextual semantics and business goals.

% 最近的进步，包括深度学习技术的整合，已经试图解决其中的一些限制。例如，深度聚类方法将表示学习与聚类任务结合起来，能够更好地处理复杂的非结构化数据。然而，这些方法的主要目标是优化集群性能，而不是评估与特定业务目标的一致性。此外，像CONNER这样的框架为知识密集型任务引入了全面的内在和外在评估指标，重点关注事实性、一致性和相关性等方面。虽然这些方法很有前途，但它们是为评估生成的知识而设计的，而不是直接评估Taxonomy的质量，这突出了为使Taxonomy与业务需求保持一致而量身定制的评估方法的差距。

Recent advancements, including the integration of deep learning techniques, have attempted to address some of these limitations~\cite{Zhou2022A}. For example, deep clustering approaches combine representation learning with clustering tasks, enabling better handling of complex, unstructured data~\cite{Wang2024An,Wang2023A,Yang2024Fuzzy-Based}. However, these methods primarily target optimizing clustering performance rather than evaluating alignment with specific task-oriented objectives. Additionally, frameworks like CONNER~\cite{Chen2023Beyond} have introduced comprehensive intrinsic and extrinsic evaluation metrics for knowledge-intensive tasks, focusing on aspects like factuality~\cite{Martínez-Plumed2015Knowledge}, coherence, and relevance~\cite{Chen2023Beyond,Cao2022On}. While promising, these methods are designed for evaluating generated knowledge rather than assessing Taxonomy quality directly, highlighting the gap in evaluation approaches tailored for Taxonomy's alignment with real world task-oriented needs.

\section{Method}\label{sec:method}
\subsection{Framework}  
The evaluation process in this paper is meticulously designed and divided into three main steps to ensure accuracy, efficiency, and fairness:

\subsubsection{Step 1: Subtree Selection}  
\label{sec:alg}
The subtree selection step aims to balance the quality and efficiency of the evaluation.  
Starting from the root node, the evaluation follows a top-down, breadth-first traversal strategy. This approach allows a gradual exploration of the tree structure, ensuring that each selected subtree maintains structural rationality. This rationality reflects the overall characteristics of the tree while avoiding excessive granularity, thus ensuring evaluation efficiency.  

To reduce computational overhead during the evaluation process, \methode~focuses on the number of nodes in subtrees. Specifically, \methode~sets a range for the size of subtrees to ensure a balance between depth and breadth in the evaluation. This range is defined based on the average out-degree \( avg.D_{out}(T) \) and the tree height \( H(T) \), with the range given by  
\begin{equation}  
    [ avg.D_{out}(T) \cdot H(T) \cdot k, \  avg.D_{out}(T) \cdot H(T) \cdot 2 \cdot k]  
    \label{eq:1}
\end{equation}  
where \( k \) is a variable parameter.  
This approach reduces computational resource consumption while ensuring that the selected subtrees are sufficiently representative and can serve as valid samples for subsequent evaluations.
The specific implementation details of the algorithm are provided in Sec.\ref{sec:apx:alg}.

\subsubsection{Step 2: Scoring}  
After subtree selection, the selected subtrees are provided to the LLM for calculating predefined evaluation metrics, which are essential for ensuring the interpretability and applicability of the results. 
To minimize biases in the evaluation, \methode~ensures that each subtree is assessed under consistent conditions. This is achieved through two main measures: standardized input formats and cross-validation techniques.

\paragraph{Standardized Input Format}  
\methode~standardizes the input by converting all subtrees into a unified JSON format for consistent parsing by the LLM. Three evaluation objectives are defined for taxonomy assessment:
1) Evaluation of the "concept" itself;
2) Evaluation of the relationship between the "parent concept" and its "child concepts"; 
3) Evaluation of the relationship between the "parent concept" and a group of "child concepts";  
Standardized input templates are designed for these three objectives, with details provided in Sec.\ref{sec:apx:input}. 

\paragraph{Cross-Validation Techniques}  
To reduce data bias in the evaluation and improve result stability, we adopt a two-stage cross-validation strategy. 
The first stage is subtree disturbance, where multiple similar subtree versions are created by randomly rearranging nodes at the same level. This helps reduce over-reliance on one structure and improves the generalization of the evaluation.
The second stage is multi-round scoring, where each set of variant subtrees is independently assessed. By conducting three rounds of evaluation and taking the average of all results as the final score, this method effectively reduces the influence of outliers caused by random factors or model biases, ensuring the reliability and consistency of the evaluation results.

\subsubsection{Step 3: Handling Extreme Subtrees}  
In the evaluation process, extreme subtree instances with unusual edge counts may arise, potentially negatively impacting the accuracy of the evaluation results. 
We find that an excessive or insufficient number of edges in a subtree can lead to biases in neighborhood-level metrics. 
To address these special cases, \methode~sets clear boundaries for the number of edges in the subtree selection phase, as defined in Equation \ref{eq:1}. If the number of edges in a subtree exceeds the upper limit or falls below the lower limit, \methode~applies penalty measures by adjusting the scoring formula to reflect its anomaly.

When the number of edges in a subtree exceeds the predefined upper limit, the penalty formula used is:
\begin{equation}
P = -\lambda \cdot \max\left(1, \frac{|cur\_subtree|}{threshold\_high}\right)
\label{eq:2}
\end{equation}
, where \( |cur\_subtree| \) represents the number of edges in the current subtree, \( threshold\_high = avg.D_{out}(T) \cdot H(T) \cdot 2k \) (the upper limit defined in Eq.\ref{eq:1}), and \( \lambda \) is a positive penalty coefficient.

Conversely, when the number of edges in a subtree falls below the predefined lower limit, the penalty formula is:
\begin{equation}  
P = -\mu \cdot \max\left(1, \frac{threshold\_low}{|cur\_subtree|}\right)   
\label{eq:3}  
\end{equation}  
, where \( |cur\_subtree| \) represents the number of edges in the current subtree, \( threshold\_low = avg.D_{out}(T) \cdot H(T) \cdot k \) (the lower limit defined in Eq.\ref{eq:1}), and \( \mu \) is another positive penalty coefficient.

\subsection{Evaluation Metrics}  
\label{sec:metric}
To ensure taxonomies align with practical business needs, we propose the following evaluation metrics:
\paragraph{Single Concept Accuracy (SCA)}:  
    Concepts should be clear, precise, and easily understood to meet practical use cases. SCA evaluates whether the concepts are unambiguous and easy to comprehend, ensuring that terms are specific enough to accurately reflect the intended meaning.  
    
\paragraph{Hierarchy Relationship Rationality (HRR)}:  
    A taxonomy’s hierarchical relationships should reflect logical and practical connections between concepts. HRR measures whether parent-child relationships are valid, fully encompassing, and free from contradictions, ensuring the hierarchy aligns with real-world business structures.

\paragraph{Hierarchy Relationship Exclusivity (HRE)}:  
    Parent concepts should be exclusive enough to clearly distinguish between related sub-concepts. HRE evaluates whether a parent concept effectively differentiates between subordinate concepts, ensuring that each concept is categorically distinct for practical use.

\paragraph{Hierarchy Relationship Independence (HRI)}:  
    Concepts should minimize overlap to avoid confusion and ensure clear categorization. HRI measures the degree of independence between concepts of a single parent, ensuring that concepts in the same hierarchical level are distinct and logically separate.

Sec.\ref{sec:apx:input} provides examples of each metric, showcasing both good and bad cases.

\section{Experiment}
\subsection{Dataset}

% 3个taxonomy的统计指标
% taxonomy的不同版本

The experiments in this section involve taxonomy datasets from multiple domains, including the open-source MAG-CS dataset in the computer science domain and the Ali-Taxo dataset used in the real-world business operations of Alibaba Group. These datasets were processed and adapted to form the following subsets:  

\begin{itemize}
    \item \textbf{MAG-CS.large}: Full dataset of MAG-CS, manually constructed and automated-expanded series of concepts, primarily focusing on computer science, aiming for comprehensive coverage.
    \item \textbf{MAG-CS.small}: Subtree starting from "Artificial Intelligence" node in MAG-CS, covering a specific portion of the computer science field.
    \item \textbf{Ali-Taxo.cluster}: Taxonomy constructed by A using automated clustering methods on internal corpus from well-known websites, resulting in a well-clustered and concept-rich structure.
    \item \textbf{Ali-Taxo.LLM}: Taxonomy built by A on Ali-Taxo.cluster using LLM, applied to the same internal corpus, aiming for richer concepts and more reasonable relationships.
    \item \textbf{Ali-Taxo.LLM.Picked}: Manually selected high-quality subset from Ali-Taxo.LLM, serving as an expert-validated evaluation benchmark for reliable reference in subsequent analysis and research.
    \item \textbf{Ali-Taxo.LLM.Rand}: Randomly rearranged nodes and relationships based on Ali-Taxo.LLM.Picked, creating a completely random tree structure for testing algorithm sensitivity to taxonomy structure changes.
    \item \textbf{Ali-Taxo.LLM.Reverse}: Reversed nodes and relationships based on Ali-Taxo.LLM.Picked, used to test the impact of directional relationships in the taxonomy structure on algorithms or analysis tasks.
    \item \textbf{Ali-Taxo.LLM.Short}: All nodes connected to the root node based on Ali-Taxo.LLM.Picked, creating a very flat structure for testing algorithm performance in extreme cases.
\end{itemize}

Specific statistics of these datasets are shown in Sec.\ref{sec:apx:dataset}.

\subsection{Baselines and Settings}

To validate the reliability of \methode, this paper introduces human expert judgments based on manual evaluation. In this baseline, domain experts annotate the knowledge base, and their scoring results serve as the gold standard to quantify the upper limit of the accuracy of the automated evaluation method. The specific annotation process is described in Sec.\ref{sec:apx:human}. We use the Pearson Correlation Coefficient to quantify the linear correlation between the evaluation results of \methode~and human expert judgments to validate the consistency between them. 

In the standard configuration of \methode, we adopt a combination of “in-context learning” and “few-shot learning”. Specifically, “in-context learning” provides detailed scoring guidelines and criteria, offering the model clear evaluation requirements and task instructions; while “few-shot learning includes 3 representative task examples, allowing the model to learn task patterns and scoring strategies through comparison and imitation. 

To further assess the effectiveness of \methode~, several comparison baselines are designed to investigate the impact of different settings on the evaluation results:

\begin{itemize}
    \item \textbf{LLM-Only}: The prompt contains only the Taxonomy, without scoring criteria and examples. The model entirely relies on its built-in knowledge and understanding to generate the score. 
    \item \textbf{\methode~-w/o-ICL} (without in-context learning): The prompt includes only examples, without providing scoring criteria and context. 
    \item \textbf{\methode~-0-shot} (zero-shot setting): The prompt does not include any example information, only providing scoring criteria.
\end{itemize}

The GPT-4o-0806(GPT-4o) used in \methode's standard settings serves as the baseline evaluation model, with powerful language understanding and generation capabilities. The deterministic parameters of GPT-4o (temperature=0.1) are specified. 
In the following experiments, the variable parameter \( k = 1 \) in Eq.\ref{eq:1}, and the penalty coefficients \( \lambda = 0.5 \) and \( \mu = 0.5 \) in Eq.\ref{eq:2} and Eq.\ref{eq:3} are set to ensure effective evaluation scores.

\subsection{Evaluation Result}

\begin{table}[]
\centering
\small
\resizebox{\linewidth}{!}{
\begin{tabular}{@{}l|lllll@{}}
\toprule
                              & Method            & SAli-r      & HRR-r      & HRE-r       & HRI-r       \\ \midrule
\multirow{4}{*}{MAG-CS.large} & LLM-Only          & 0.58          & 0.78          & 0.11          & 0.43          \\
                              & \methode -w/o-ICL & 0.69          & 0.83          & 0.80          & 0.85          \\
                              & \methode -0-shot  & 0.53          & 0.88          & 0.50          & 0.63          \\
                              & \methode          & \textbf{0.77} & \textbf{0.90} & \textbf{0.83} & \textbf{0.88} \\ \midrule
Ali-Taxo.cluster               & LLM-Only          & \textbf{0.79} & 0.72          & 0.58          & -0.12         \\
                              & \methode -w/o-ICL & 0.73          & 0.82          & 0.87          & 0.80          \\
                              & \methode -0-shot  & 0.71          & 0.82          & 0.70          & 0.34          \\
                              & \methode          & 0.77          & \textbf{0.84} & \textbf{0.88}          & \textbf{0.81} \\ \midrule
Ali-Taxo.LLM                   & LLM-Only          & \textbf{0.82} & 0.83          & 0.41          & 0.28          \\
                              & \methode-w/o-ICL  & 0.68          & 0.87          & 0.72          & 0.79          \\
                              & \methode -0-shot  & 0.68          & 0.82          & 0.78          & 0.73          \\
                              & \methode          & 0.76          & \textbf{0.92} & \textbf{0.83} & \textbf{0.83} \\ \bottomrule
\end{tabular}
}

\caption{Comparison of LLM baselines, where SAli-r, HRR-r, HRE-r, and HRI-r represent the correlations of the four metrics with human expert annotations.}
\label{tab:baseline}
\end{table}
\begin{table}[]
\centering
\small
\resizebox{\linewidth}{!}{
\begin{tabular}{@{}clllll@{}}
\toprule
\multicolumn{1}{c|}{}                              & Method                    & SCA  & HRR  & HRE  & HRI  \\ \midrule
\multicolumn{6}{c}{\cellcolor[HTML]{C0C0C0}MAG-CS}                                                         \\ \midrule
\multicolumn{1}{c|}{MAG-CS.large}                  & LLM-Only                  & 8.53 & 6.19 & 3.67 & 3.63 \\
\multicolumn{1}{c|}{}                              & \methode-w/o-ICL          & 7.96 & 5.11 & 6.33 & 5.71 \\
\multicolumn{1}{c|}{}                              & \methode -0-shot          & 8.57 & 6.89 & 7.31 & 6.49 \\
\multicolumn{1}{c|}{}                              & \methode                  & 8.11 & 5.68 & 6.21 & 5.54 \\ \midrule
\multicolumn{1}{c|}{\multirow{2}{*}{MAG-CS.small}} & \methode                  & 8.09 & 5.10 & 5.83 & 5.82 \\
\multicolumn{1}{c|}{}                              & \methode *                & 8.15 & 4.98 & 5.57 & 5.79 \\ \midrule
\multicolumn{6}{c}{\cellcolor[HTML]{C0C0C0}Ali-Taxo}                                                        \\ \midrule
\multicolumn{1}{c|}{Ali-Taxo.cluster}               & LLM-Only                  & 7.60 & 5.96 & 7.24 & 7.99 \\
\multicolumn{1}{c|}{}                              & \methode -w/o-ICL         & 7.96 & 7.40 & 6.25 & 5.83 \\
\multicolumn{1}{c|}{}                              & \methode -0-shot          & 5.89 & 7.41 & 5.02 & 6.51 \\
\multicolumn{1}{c|}{}                              & \methode                  & 6.84 & 7.02 & 5.42 & 4.64 \\ \midrule
\multicolumn{1}{c|}{Ali-Taxo.LLM}                   & LLM-Only                  & 6.83 & 3.99 & 2.28 & 3.66 \\
\multicolumn{1}{c|}{}                              & \methode-w/o-ICL          & 8.67 & 4.06 & 4.29 & 6.18 \\
\multicolumn{1}{c|}{}                              & \methode-0-shot           & 7.74 & 4.35 & 5.00 & 7.33 \\
\multicolumn{1}{c|}{}                              & \methode                  & 7.34 & 4.01 & 5.34 & 6.42 \\ \midrule
\multicolumn{1}{c|}{Ali-Taxo.LLM.Picked}            & \multirow{4}{*}{\methode} & 9.03 & 5.94 & 5.68 & 6.35 \\
\multicolumn{1}{c|}{Ali-Taxo.LLM.Rand}              &                           & 9.03 & 1.60 & 1.25 & 5.41 \\
\multicolumn{1}{c|}{Ali-Taxo.LLM.Reverse}           &                           & 9.03 & 0.12 & 0.97 & 5.84 \\
\multicolumn{1}{c|}{Ali-Taxo.LLM.Short}             &                           & 9.03 & 6.18 & 0.00 & 6.07 \\ \bottomrule
\end{tabular}
}
\caption{The evaluation results of baselines on different Taxonomies. The \methode* for MAG-CS.small refers to the results evaluated using the \methode method on the subset of MAG-CS.small corresponding to MAG-CS.large.}
\label{tab:result}
\end{table}

The results of different baselines are shown in Tab.\ref{tab:baseline}. Meanwhile, Tab.\ref{tab:result} presents the evaluation results of these methods on different taxonomies for each metric. Based on these experimental data, we have the following conclusions:

\textbf{\methode~demonstrates high reliability in complex evaluation tasks. } 
In tasks involving multi-level semantic reasoning, \methode~shows evaluation capabilities close to those of human experts. Especially in the HRR and HRE metrics, which require deep semantic understanding, \methode~achieved Pearson correlation coefficients of 0.90 and 0.83 with human annotations, significantly outperforming baseline methods that rely solely on pre-trained model outputs, with correlation coefficients of 0.78 and 0.11, respectively. It also shows its advantages in structural anomaly detection (HRI).  
For example, when evaluating the knowledge base Ali-Taxo.LLM.Reverse, where the relationship direction is inverted, \methode's HRR score was only 0.12, with a very small error compared to the human score of 0.15. This demonstrates \methode's ability to accurately identify logical conflicts.  
Furthermore, although the LLM-Only baseline performs better on simpler semantic accuracy metrics like SCA (e.g., SCA for MAG-CS.large is 8.53), it exhibits systematic bias in more complex tasks. For instance, in Ali-Taxo.LLM.Picked, the HRI score was 3.66, significantly lower than the human score of 6.42. This reflects that LLM tends to misinterpret "parent nodes with rich child nodes" as "rich context," exposing limitations in autonomous reasoning. This suggests that incorporating domain-specific constraints is essential.

\textbf{Contextual example is the core mechanism to enhance evaluation consistency.  }
Experimental data show that removing example information from \methode~(i.e., \methode~-0-shot) leads to a significant drop in the average correlation for the exclusivity metric, with HRE-r for MAG-CS.large decreasing from 0.83 to 0.50. In contrast, with examples retained but without context, HRE-r stabilizes at 0.80. This indicates that concrete scenario demonstrations effectively unify scoring standards and prevent the model from over-generalizing. For example, when evaluating the relationship "tool → 3D modeling software," \methode~-0-shot, without examples, often considers edge cases as reasonable. However, with complete example-based learning, the method accurately identifies the semantic range shift between the two terms.

\textbf{\methode~demonstrates stability in evaluations across different databases.  
}Table \ref{tab:result} shows the evaluation results for both MAG-CS.large and MAG-CS.small knowledge bases. For MAG-CS.small, the evaluation results from the full taxonomy and its corresponding subset show a high correlation coefficient of 0.92, with only a 4.5\% difference in the HRE metric scores. This indicates that the subtree selection strategy did not compromise the semantic integrity of the taxonomy, allowing \methode~to maintain stable evaluation performance across different data scales.

\textbf{\methode~is highly sensitive to taxonomy structural anomalies.  }
\methode~effectively detects various structural defects. For example, in Ali-Taxo.LLM.Rand, randomizing relationships caused the HRR score to drop to 1.60, compared to the original structure's score of 5.94. In Ali-Taxo.LLM.Reverse, inverting relationship directions led to an HRR score close to zero. In Ali-Taxo.LLM.Short, the flattened structure severely damaged HRE, with the score dropping to zero. These results, with errors within 5\% compared to human evaluation, fully demonstrate \methode's effectiveness in monitoring data corruption (e.g., relationship misplacement) and modeling errors (e.g., hierarchical flattening).

\section{Analysis}
\subsection{Validity Analysis}
\subsubsection{Efficiency Analysis}
% 评估效率
% 画出Correlation、Token数随着子树中Node变化的曲线

\begin{figure}
    \centering
    \includegraphics[width=0.45\textwidth]{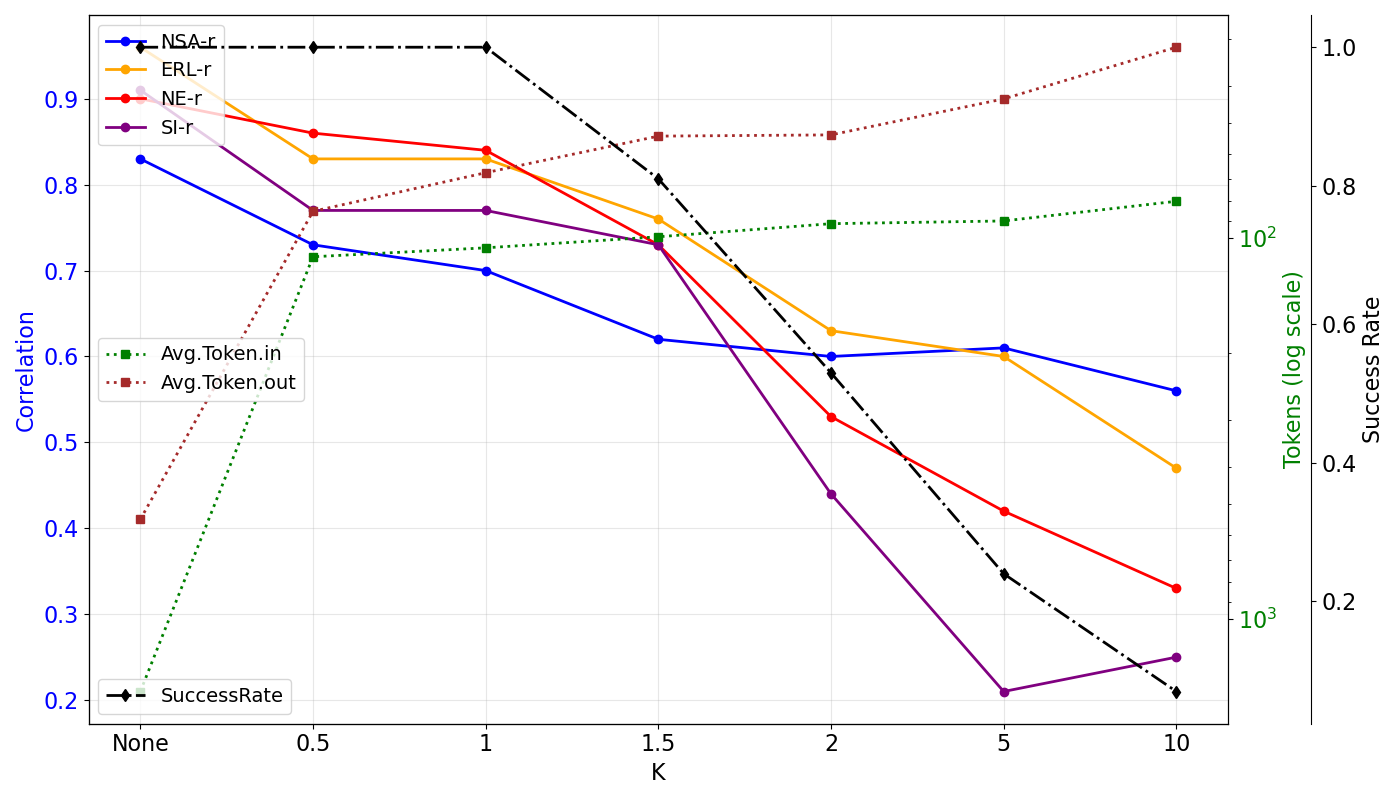}
    \caption{The trend of evaluation efficiency with subtree node count, where the solid line shows the success rate, and the dashed lines represent the average token cost for input (black), output (red), and input (green). The token cost axis is logarithmic, with higher values (lower position) indicating lower efficiency. The parameter K in formula \ref{eq:1} defines the subtree size, with K set to None indicating evaluation with the smallest subtree.}
    \label{fig:efficiency}
\end{figure}

Figure \ref{fig:efficiency} illustrates the trend of evaluation efficiency as the number of subtree nodes varies. When \( K \) is set to None, the evaluation uses the minimal subtree consisting of a single node, relation, or domain. As \( K \) increases, the size of the subtree input to the LLM expands accordingly. Results show that when subtree settings are enabled, compared to direct input, the average input and output token costs decrease from 1549.10 and 547.40 to 106.08 and 67.38, representing reductions of 93.15\% and 87.69\%, respectively. A fine-grained data analysis leads to the following key conclusions:  

\textbf{A trade-off between cost and accuracy is necessary to choose the optimal parameters.}  
As subtree size increases, the evaluation success rate declines and approaches zero with further expansion, indicating that input scale must be carefully considered. When K increases from the smallest range (atomic operations) to 0.5, input and output token counts decrease significantly, with only a slight decrease in the average correlation coefficient, suggesting that moderate subtree expansion reduces cost while maintaining quality. However, when K increases from 1 to 1.5, the correlation coefficient drops substantially, and the success rate declines from 100\% to 81\%, indicating that the benefits of expanding the subtree level off when K is around 1.

\textbf{The LLM's bottleneck in handling long contexts significantly impacts evaluation.}  
LLMs experience attention dilution, where long inputs obscure critical information. Experiments show that with K set to 5, the correlation coefficient drops to 0.21, whereas with the smallest range evaluation, it increases to 0.91. Long texts may also cause reasoning chain breaks, linked to the LLM’s long-context forgetting. For example, when K reaches 2, performance drops, with a 43\% missed detection rate for relationships in the latter part of the subtree. When K exceeds 1.5, the success rate decreases non-linearly from 81\% to 53\% to 24\%, due to the model's limitations in handling large token amounts.

\begin{table*}[]
\centering
\small
\caption{\methode's evaluation results by different models.}
{
\begin{tabular}{@{}l|llllllll@{}}
\toprule
Model     & SCA  & SCA-r & HRR & HRR-r & HRE & HRE-r  & HRI & HRI-r \\ \midrule
Qwen2-7b  & 9.31 & -0.03         & 7.88 & -0.10         & 8.13 & 0.21          & 8.98 & 0.35          \\
Qwen2-72b & 8.40 & 0.69          & 6.66 & 0.65          & 6.23 & 0.38          & 7.96 & 0.32          \\
GPT-4o    & 8.09 & \textbf{0.70} & 5.10 & \textbf{0.83} & 5.83 & \textbf{0.84} & 5.82 & \textbf{0.77} \\
\bottomrule
\end{tabular}
}

\label{tab:ablation}
\end{table*}

\subsubsection{Model Analysis}

Table \ref{tab:ablation} presents the evaluation results of MAG-CS.small using different models. Specifically, the Qwen2-7b model performs poorly across all metrics, particularly in terms of correlation with human evaluations, where it often shows negative or low correlation. This suggests its understanding of the taxomony is vague, and it tends to give higher scores to concepts that are less certain. In contrast, the Qwen2-72b model shows improvements in correlation for NSA and HRR, especially in NSA-r and HRR-r, indicating its evaluations in these areas are closer to human assessments. However, for other metrics like HRE and HRI, Qwen2-72b still underperforms compared to GPT-4o. 
GPT-4o outperforms all models across all evaluation metrics, especially in HRR, HRE, and HRI, with higher correlations to human evaluations, reaching 0.83, 0.84, and 0.77, respectively. This demonstrates its strong capabilities in complex semantic understanding and reasoning, clearly surpassing the Qwen series models. 
These results show that stronger models, like GPT-4o, are more stringent in taxomony evaluations and can more accurately understand and assess concepts, while weaker models, like Qwen2-7b, tend to give higher scores to concepts that are uncertain or vague. 
Therefore, the capability of the evaluation model has a significant impact on the evaluation results.

\subsubsection{Metric Analysis}
% Metrics的设置

Tab.\ref{tab:case} presents typical examples demonstrating \methode's evaluation capabilities across four dimensions. 

In SCA, \methode~accurately identifies terminology issues, such as correcting the misspelled ``prothesis'' to ``prosthesis'', showcasing its strict validation ability for technical terms.
In HRR evaluation, \methode~effectively distinguishes between reasonable and unreasonable associations. For instance, it evaluates the relationship ``artificial intelligence → overscan'' as weakly correlated (HRR=1), while ``artificial intelligence → description logic'' is assessed as a strong logical relation (HRR=10), ensuring the accuracy of relationships in the taxomony.
For HRE, \methode~precisely identifies the strong correlation between ``map projection'' and its neighboring terms (HRE=10), while also recognizing the semantic confusion in the neighborhood of ``fuzzy logic'' (HRE=3), demonstrating its sharp judgment in semantic aggregation.
In HRI evaluation, \methode~detects redundant associations, such as the redundant neighboring nodes of ``fuzzy logic'' (HRI=5), and evaluates the clarity of the neighboring node structure of ``map projection'' (HRI=9), significantly enhancing the rationality of knowledge organization.

These results demonstrate that \methode's evaluation system can systematically identify semantic errors, logical contradictions, and structural defects in a Taxonomy. Its multi-dimensional scoring mechanism is highly consistent with human expert judgment, providing quantifiable and actionable directions for optimizing the quality of taxomonies.

\begin{table*}
\caption{Cases demonstrating \methode's evaluation capabilities across 4 metrics, including scores and reasons.}

\small
\resizebox{0.98\textwidth}{!}{
\begin{tabular}{@{}p{1.5cm}p{4.5cm}p{2cm}p{6.5cm}@{}}
\toprule
\multicolumn{1}{l}{\textbf{Metrics}} & \textbf{Case}                                                                                                                                                                                                                                                          & \textbf{Score}                                                   & \textbf{Reason}                                                                                                                                                                                                                                                                                                                                                                                                                                      \\ \midrule
\multirow{2}{*}{SCA}              & prothesis                                                                                                                                                                                                                                                            & 0                                                    & Spelling error, maybe it should be 'prosthesis'.                                                                                                                                                                                                                                                                                                                                                        \\ \cmidrule(l){2-4} 
                                  & prosthesis                                                                                                                                                                                                                                                           & 10                                                   & It is broadly applicable to any device used to replace a part of the human body or its functions, and the definition is concise and clear.                                                                                                                                                                                                                                              \\ \midrule
HRR                               & [artificial intelligence, overscan]                                                                                                                                                                                                                         & 1                                                    & Overscan is a concept in display technology and has no direct connection with artificial intelligence.                                                                                                                                                                                                                                                                                              \\ \cmidrule(l){2-4} 
\multicolumn{1}{l}{}              & [artificial intelligence, description logic]                                                                                                                                                                                                                & 10                                                   & Description logic is an important tool for knowledge representation and reasoning, and is closely related to artificial intelligence.                                                                     \\ \midrule
{HRE \& HRI}          & map projection, [lambert azimuthal equal area projection, albers equal area conic projection, polyconic projection, web mercator, equidistant conic projection]                                                                                             & HRE=10, HRI=9 & Map projection' clearly distinguishes between map projections and other types, and there are clear differences between the hyponyms and hyponyms, each with a specific purpose.                                                                                                                                             \\ \cmidrule(l){2-4} 
                                  & fuzzy logic, [chu space,truth function, fuzzy extractor,fril,iris flower data set,adaptive resonance theory,soft computing,cerebellar model articulation controller,fuzzy clustering,granular computing, vikor method,modus ponens, fuzzy set,(...21 more)] & HRE=3, HRI=5 & The hypernym 'fuzzy logic' only defines the hyponyms to some extent, but many of the hyponyms involve concepts beyond the direct scope of fuzzy logic, such as 'iris flower data set' and 'cue space'. Although many hyponyms involve different sub-fields, they often overlap in fuzzy theory applications or related research. \\ \bottomrule
\end{tabular}
}

\label{tab:case}
\end{table*}

\subsection{Taxonomy Analysis} 
\label{ana:taxo}
This section analyzes the evaluation results of different taxomonies using \methode~and highlights their inherent issues.

\subsubsection{MAG-CS}  
MAG-CS, an automatically constructed academic resource, exhibits three main issues:

\paragraph{Terminology Standardization Issues}  
MAG-CS shows significant terminology standardization problems due to its automated extraction, with 12.7\% of nodes containing undefined abbreviations, leading to confusion during retrieval. This underscores the need for stricter standardization in future taxomony construction.

\paragraph{Shallow Relationship Semantics}  
MAG-CS places a significant emphasis on surface-level semantics, which unfortunately leads to the formation of inappropriate associations. 
A notable example is the linkage between “Diplomatic Protocol” and “Network Protocol,” a connection established solely on the basis of lexical similarity rather than meaningful conceptual alignment.
This highlights the limitations of automated methods in semantic understanding.

\paragraph{Hierarchical Structure Collapse}  
The hierarchical structure in MAG-CS is characterized by an instability that arises from the improper placement of concepts, such as the juxtaposition of the broad term ``algorithm'' with the more specific ``difference map algorithm''. This affects logical consistency and user navigation, emphasizing the challenges of automated methods in concept organization.

\subsubsection{CA-Taxo.cluster}  
CA-Taxo.cluster, built using automatic text clustering techniques, reveals the following issues:

\paragraph{Semantic Ambiguity}  
Many concepts in CA-Taxo.cluster lack precise definitions, such as the forced combination of ``Corporate Financial Reports and Feng Shui Culture''. This reduces usability and exposes limitations in clustering techniques.

\paragraph{Poor Relationship Quality}  
CA-Taxo.cluster has many nodes linked to unrelated concepts due to reliance on statistical co-occurrence. For example ``Supply Chain Management'' and ``Five Elements Theory''. This leads to misleading associations and highlighting the need for better relationship accuracy through domain knowledge and logical reasoning.

\subsubsection{CA-Taxo.LLM}  
CA-Taxo.LLM built by advanced LLM presents new quality issues affecting user experience:

\paragraph{Semantic Hollow Nodes}  
CA-Taxo.LLM has 24.1\% zero-outdegree nodes, like ``elevator pitch''. These nodes lack meaningful content or connections. This reflects a tendency of LLMs. They prioritize long-tail coverage. This can be at the expense of quality. It may potentially confuse users.

\paragraph{Imbalanced Concept Granularity}  
The taxonomy shows imbalances in granularity. Specifically, it presents a scenario where broad, overarching topics are found alongside ultra-specific, highly detailed nodes, all existing at the same hierarchical level. Broad topics and ultra-specific nodes coexisting at the same level, such as ``Service'' alongside ``Bus 181 stop coordinates'', disrupting logical structure.

\paragraph{Hallucination Interference}  
Hallucinations, such as ``Quantum Entanglement-Based Customer Satisfaction Prediction'', are a critical issue, as LLMs prioritize linguistic coherence over factual accuracy. This undermines trust in the taxomony, emphasizing the need for verification mechanisms.

\section{Conclusion}
We present \methode~, an innovative taxonomy evaluation method that leverages the power of LLMs to achieve efficient, flexible, and accurate assessment. 
It tackles the challenges of efficiency, fairness, and consistency through hierarchical evaluation, cross-validation, and standardized input formats, while offering comprehensive results through multidimensional evaluation metrics. 
The experimental performance of \methode~has been exemplary, demonstrating its ability to effectively pinpoint defects within taxonomies. 
In summary, \methode~stands as a testament to the potential of integrating LLMs into the field of knowledge evaluation.

\bibliography{custom}

\newpage
\section{Appendix}
\label{sec:appendix}
\subsection{Subtree Selection Algorithm}
\label{sec:apx:alg}
Alg.\ref{alg:select} shows the complete algorithm in Sec.\ref{sec:alg}.

\begin{algorithm}
\caption{Subtree Selection Algorithm}
\label{alg:select}
\begin{algorithmic}[1]
\State \textbf{Notation:} $n$: Node, $R(n, c)$: Relations between $n$ and child $c$, $R_{max}$: Max relations per subtree, $subtree\_list$: List of selected subtrees, $cur\_tree$: Current subtree, $Q$: BFS queue, $G$: Subgroups of $R(n, c)$
\State \textbf{Input:} $R_{max}$
\State \textbf{Output:} $subtree\_list$
\State \textbf{Initialize:} $subtree\_list$, $cur\_tree$, $Q$ 
\State Enqueue the root node to $Q$
\While{$Q$ not empty}
    \State $n \gets$ Dequeue from $Q$
    \For{each child $c$ of $n$}
        \If{$\lvert R(n, c) \rvert + \lvert cur\_tree \rvert \leq R_{max}$}
            \State Add $R(n, c)$ to $cur\_tree$
        \Else
            \State Add $cur\_tree$ to $subtree\_list$, Clear $cur\_tree$
            \If{$\lvert R(n, c) \rvert > R_{max}$}
                \State $G \gets \text{split}(R(n, c), R_{max})$
                \For{each $g \in G$}
                    \State Add $g$ to $subtree\_list$
                \EndFor
                \State Add remaining $R(n, c)$ to $cur\_tree$
            \Else
                \State Add $R(n, c)$ to $cur\_tree$
            \EndIf
        \EndIf
        \State Enqueue $c$ to $Q$
    \EndFor
\EndWhile
\State Add any remaining $cur\_tree$ to $subtree\_list$
\State \textbf{End}
\end{algorithmic}
\end{algorithm}

\subsection{Input Format And Metric Cases}
\label{sec:apx:input}
\begin{table*}[]
    \centering
\begin{tabular}{@{}c@{}}
\begin{subtable}{\linewidth}
\centering
\caption{Concept Evaluation}
\resizebox{0.99\textwidth}{!}{
\begin{tabular}{@{}p{2.3cm}p{20cm}@{}}
\toprule
Evaluation Objective         & Concept-Level Evaluation \\ \midrule  
Evaluation Content Type      & Single Concept \\ \midrule  
Evaluation Metric            & Single Concept Accuracy (SCA) \\ \midrule  
Evaluation Metric Definition & Evaluates the clarity and uniqueness of the concept. \\ \midrule  
Detailed Description         & The concept's semantics should be unambiguous and clearly defined, enabling users to quickly understand and classify information. If the concept's semantics are not well-defined or contain significant ambiguity and contradictions, SCA is scored as 0. If the definition is generally clear but may cause confusion in certain cases, or if the concept is slightly verbose, SCA is scored as 5. If the definition is precise, specific, free of ambiguity, and concisely expressed, SCA is scored as 10. \\ \midrule  

\multirow{3}{*}{Examples}                                       & Example 1: \{"Disease": 10, "Reason": "The concept is clear and unique, with no ambiguity."\} \\  
& Example 2: \{"A Certain Disease": 5, "Reason": "The concept is relatively clear but has some generalization, potentially referring to multiple diseases."\} \\  
& Example 3: \{"Certain": 0, "Reason": "The meaning represented by this concept is too vague to determine its specific significance."\}                                                                                                                                                           \\ \midrule
\begin{tabular}[c]{@{}l@{}}Output Format\\ (JSON)\end{tabular}   & {[}\{"Concept1":score, "Reason": "Scoring reasons and reasoning process"\},  \{"Concept2":score, "Reason": "Scoring reasons and reasoning process"\}, ..., \{"ConceptN":score, "Reason": "Scoring reasons and reasoning process"\}{]}                                    \\ \midrule
\begin{tabular}[c]{@{}l@{}}Content Format\\ (JSON)\end{tabular} & {[}"Concept1", "Concept2", ...,"ConceptN"{]}                                                                                                                                                                            \\ \bottomrule
\end{tabular}
}
\end{subtable} \\

\begin{subtable}{\linewidth}
\centering
\caption{Relation Evaluation}
\resizebox{0.99\textwidth}{!}{
\begin{tabular}{@{}p{2.3cm}p{20cm}@{}}
\toprule
Evaluation Objective         & Relation-Level Evaluation      \\ \midrule
Evaluation Content Type      & Single Relationship            \\ \midrule
Evaluation Metric            & Hierarchy Relationship Rationality (HRR)    \\ \midrule
Evaluation Metric Definition & Evaluate whether the relationships represented by edges conform to domain common sense or logical constraints.   \\ \midrule
Detailed Description         & The relationship should be logically coherent, correctly generalizable from instances, and free from contradictions or circular dependencies.  If the relationship represented by the edge clearly contradicts domain knowledge or logical constraints, or if its definition is vague, the HRR score is 0.  If the relationship generally aligns with domain knowledge or logical constraints but may have logical loopholes or cause confusion in certain cases, the HRR score is 5.  If the relationship is well-defined, specific, fully compliant with domain knowledge and logical constraints, and expressed concisely without ambiguity, the HRR score is 10.\\ \midrule
\multirow{3}{*}{Examples}                                       & Example 1: \{"Product, Digital Product": 10, "Reason": "The relationship is clear, specific, and aligns with common knowledge and logic."\} \\  
& Example 2: \{"Phone, Apple": 5, "Reason": "The term 'Apple' can also refer to a fruit, which may cause ambiguity."\} \\  
& Example 3: \{"Headache, Cancer": 0, "Reason": "This contradicts medical knowledge, resulting in a logic score of 0."\}                                                                                                                                                                                       \\ \midrule
\begin{tabular}[c]{@{}l@{}}Output Format\\ (JSON)\end{tabular}   & {[}\{"Father":father, "Child":child1,"Score": score,"Reason": "Scoring reasons and reasoning process"\}, ..., \{"Father":father, "Child":childN,"Score": score,"Reason": "Scoring reasons and reasoning process"\}{]}                                   \\ \midrule
\begin{tabular}[c]{@{}l@{}}Content Format\\ (JSON)\end{tabular} & {[}\{"Father":father, "Child":child1\}, ...\{"Father":father, "Child":childN\}{]}                                                                                                        \\ \bottomrule
\end{tabular}
}
\end{subtable} \\
\end{tabular}
    \caption{Input Content and Format for Different Evaluation Metrics.}

\label{tab:prompt2}
\end{table*}

\begin{table*}[]
    \centering
\begin{tabular}{@{}c@{}}

\begin{subtable}{\linewidth}
\centering
\caption{Hierarchy-Level Evaluation--HRE}
\resizebox{0.99\textwidth}{!}{
\begin{tabular}{@{}p{2.3cm}p{20cm}@{}}
\toprule
Evaluation Objective         & Hierarchy-Level Evaluation    \\ \midrule
Evaluation Content Type      & Node and its child nodes      \\ \midrule
Evaluation Metric            & Hierarchy Relationship Exclusivity(HRE)   \\ \midrule
Evaluation Metric Definition & Evaluate the semantic distinguishability of a parent node for its set of child nodes.                                      \\ \midrule
Detailed Description         & Evaluate the semantic cohesion of a central node's direct neighbor set, ensuring all neighboring nodes are closely related to the central node without interference. If the neighbor set contains a large number of semantically unrelated or conflicting nodes, making the central node's semantic scope unclear, HRE is 0. If the majority of the neighbor set is semantically consistent but includes some (\textless{}=50\%) peripheral noise or overly broad categories, HRE is 5. If the neighbor set is highly focused, with all nodes strictly aligned with the core semantics of the central node, HRE is 10.\\ \midrule
\multirow{3}{*}{Examples}                                                      & Example 1: \{"Convolutional Neural Network → {[}Convolutional Layer{]}, {[}Pooling Layer{]}, {[}Activation Function{]}, {[}Backpropagation Algorithm{]}": 10, "Reason": "All neighbors describe model structure or training mechanisms, with a clear semantic boundary."\}  
\\
& Example 2: \{"CNN → {[}Convolutional Layer{]}, {[}Pooling Layer{]}, {[}News Media Organization{]}": 5, "Reason": "The first two belong to neural networks, but 'News Media Organization' falls outside this scope due to the ambiguity of 'CNN'."\}  
\\
& Example 3: \{"Object → {[}Cosmetic Products{]}, {[}Amoxicillin{]}, {[}Orange{]}": 0, "Reason": "The neighbor set is highly dispersed in meaning, caused by the overly broad scope of 'Object'."\}                                                                                                     \\ \midrule
\begin{tabular}[c]{@{}l@{}}Output Format\\ (JSON)\end{tabular}   & 
{\{}"Father”:father1, “Children”: {[}child1.1, child1.2, …, child1.N{]}{\}}, 
"Score": score,
"Reason": "Scoring reasons and reasoning process"{\}} 
\\ \midrule
\begin{tabular}[c]{@{}l@{}}Content Format\\ (JSON)\end{tabular} & 
{\{}"Father”:father1, “Children”: {[}child1.1, child1.2, …, child1.N{]}{\}}
\\ \bottomrule
\end{tabular}
}
\end{subtable}\\

\begin{subtable}{\linewidth}
\centering

\caption{Hierarchy-Level Evaluation--HRI}
\resizebox{0.99\textwidth}{!}{
\begin{tabular}{@{}p{2.3cm}p{20cm}@{}}
\toprule
Evaluation Objective         & Hierarchy-Level Evaluation       \\ \midrule
Evaluation Content Type      & Node and its child nodes         \\ \midrule
Evaluation Metric            & Hierarchy Relationship Independence(HRI)    \\ \midrule
Evaluation Metric Definition & Evaluate the independence of the node and its child nodes in the local structure.            \\ \midrule
Detailed Description         & The node should be conceptually semantic and independent in terms of scope from its associated nodes, avoiding redundant or overly coupled associations. If all concepts in the hyponym list are nearly identical or indistinguishable in practical applications, the HRI score is 0; if the concepts in the hyponym list are partially unique, with some overlap (\textless{}=50\%), the HRI score is 5; if each concept in the hyponym list is completely unique, with no overlap with other concepts in the list, the HRI score is 10.                                                   \\ \midrule
\multirow{3}{*}{Examples}                                       & Example 1: \{"Convolutional Neural Network → {[}Convolutional Layer{]}, {[}Pooling Layer{]}, {[}Activation Function{]}, {[}Includes Backpropagation Algorithm{]}" : 10, "Reason": "Neighboring concepts are independent of each other, with no overlap or redundancy"\} \\  
& Example 2: \{"Fruit → {[}Apple{]}, {[}Orange{]}, {[}Citrus Fruits{]}" : 5, "Reason": "Citrus Fruits overlaps with Orange in the neighboring concepts, creating coupling."\} \\  
& Example 3: \{"Medicine → {[}Medicine{]}, {[}Cold Medicine{]}, {[}Cough Medicine{]}" : 0, "Reason": "Concepts in the neighboring set are highly coupled, all mutually contained."\}                                                                                                                           \\ \midrule
\begin{tabular}[c]{@{}l@{}}Output Format\\ (JSON)\end{tabular}   & 
{\{}"Father”:father1, “Children”: {[}child1.1, child1.2, …, child1.N{]}{\}}, 
"Score": score,
"Reason": "Scoring reasons and reasoning process"{\}} 
\\ \midrule
\begin{tabular}[c]{@{}l@{}}Content Format\\ (JSON)\end{tabular} & 
{\{}"Father”:father1, “Children”: {[}child1.1, child1.2, …, child1.N{]}{\}}
\\ \bottomrule
\end{tabular}
}
\end{subtable}\\

\end{tabular}
    \caption{Input Content and Format for Different Evaluation Metrics(2).}

\end{table*}

\begin{figure*}
    \centering
    \includegraphics[width=0.98\linewidth]{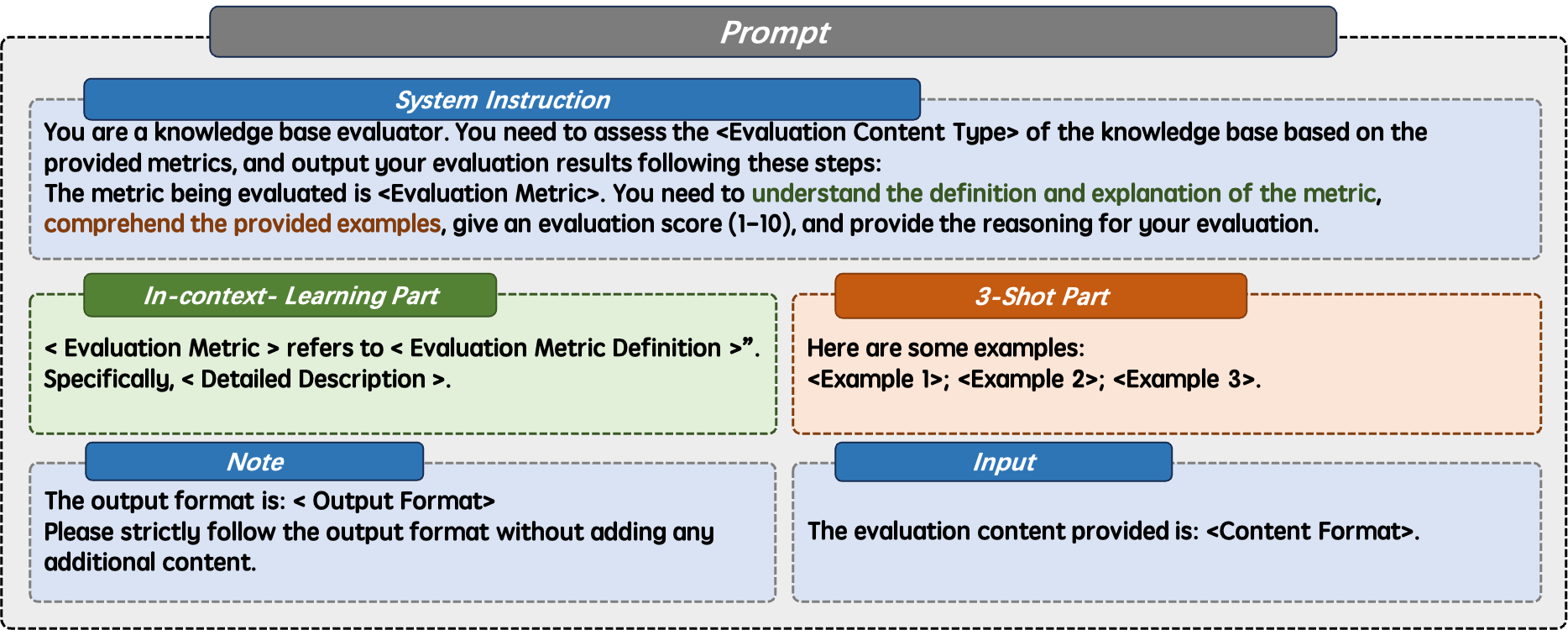}
    \caption{Evaluation prompt}
    \label{fig:prompt}
\end{figure*}

Fig.\ref{fig:prompt} shows the input format in the evaluation prompt, and \ref{tab:prompt2} shows the input content and format for different evaluation metrics by Fig.\ref{fig:prompt}.

\subsection{Dataset Details}
\label{sec:apx:dataset}

% \begin{table*}[]
% \caption{}
% \label{tab:dataset}
% \centering
% \begin{tabular}{@{}lllllll@{}}
% \toprule
% \multicolumn{1}{l|}{} & Height & \#Node & \#Relation & Max.dgree & Min.dgree & Avg.dgree \\ \midrule
% \multicolumn{7}{c}{MAG-CS}                           \\ \midrule
% \multicolumn{1}{l|}{MAG-CS.large}     & 12 & 29484 & 29483 & 1712 & 1 & 6.74 \\
% \multicolumn{1}{l|}{MAG-CS.small}     & 6  & 5988  & 5987  & 1712 & 1 & 5.94 \\ \midrule
% \multicolumn{7}{c}{AliTaxo}                             \\ \midrule
% \multicolumn{1}{l|}{AliTaxo.cluster}     &    &       &       &      &   &      \\
% \multicolumn{1}{l|}{AliTaxo.LLM}         &    &       &       &      &   &      \\
% \multicolumn{1}{l|}{AliTaxo.LLM.Picked}  &    &       &       &      &   &      \\
% \multicolumn{1}{l|}{AliTaxo.LLM.Rand}    &    &       &       &      &   &      \\
% \multicolumn{1}{l|}{AliTaxo.LLM.Reverse} &    &       &       &      &   &      \\
% \multicolumn{1}{l|}{AliTaxo.LLM.Short}   &    &       &       &      &   &      \\ \bottomrule
% \end{tabular}
% \end{table*}

\begin{table*}[]
\begin{tabular}{@{}lllllll@{}}
\toprule
\multicolumn{1}{l|}{}                    & Height & \#Node & \#Relation & Max.dgree & \begin{tabular}[c]{@{}l@{}}Min.dgree\\ (w/o Leef)\end{tabular} & \begin{tabular}[c]{@{}l@{}}Avg.dgree\\ (w/o Leef)\end{tabular} \\ \midrule
\multicolumn{7}{c}{\cellcolor[HTML]{C0C0C0}MAG-CS}                                                                                                                                                                    \\ \midrule
\multicolumn{1}{l|}{MAG-CS.large}        & 12     & 29,484 & 29,483     & 1712      & 1                                                              & 6.74                                                           \\
\multicolumn{1}{l|}{MAG-CS.small}        & 6      & 5,988  & 5,987      & 1, 712    & 1                                                              & 5.94                                                           \\ \midrule
\multicolumn{7}{c}{\cellcolor[HTML]{C0C0C0}Ali-Taxo}                                                                                                                                                                   \\ \midrule
\multicolumn{1}{l|}{Ali-Taxo.cluster}     & 4      & 2,548  & 2547       & 222       & 6                                                              & 6.53                                                           \\
\multicolumn{1}{l|}{Ali-Taxo.LLM}         & 4      & 28,117 & 28,116     & 560       & 3                                                              & 7.44                                                           \\
\multicolumn{1}{l|}{Ali-Taxo.LLM.Picked}  & 4      & 201    & 200        & 32        & 1                                                              & 5.32                                                           \\
\multicolumn{1}{l|}{Ali-Taxo.LLM.Rand}    & 4      & 201    & 200        & 24        & 1                                                              & 3.77                                                           \\
\multicolumn{1}{l|}{Ali-Taxo.LLM.Reverse} & 4      & 201    & 200        & 1         & 1                                                              & 1                                                              \\
\multicolumn{1}{l|}{Ali-Taxo.LLM.Short}   & 2      & 201    & 200        & 200       & 200                                                            & 200                                                            \\ \bottomrule
\end{tabular}
\caption{Dataset details}
\label{tab:dataset}
\end{table*}
Tab.\ref{tab:dataset} shows the statistical information of taxonomies in our experiments.

\subsection{Human Export Annodation}
\label{sec:apx:human}

To validate the reliability of \methode, this paper designs a Human expert annotation based on manual evaluation, whose scoring results will serve as the gold standard to quantify the upper limit of accuracy for automated evaluation methods. The annotation team consists of three graduate students in computer science, all with experience in the construction and management of Taxonomy. To ensure the quality of annotation, the team members first systematically studied the definitions and examples of the four evaluation metrics (see Section \ref{sec:metric}), and clarified the scoring details for dimensions such as node semantics, edge relationships, and neighborhood. In addition, through a pre-annotation test, the team further calibrated their understanding of domain concepts, ensuring consistency in understanding.
The annotation process is divided into three stages. Firstly, each annotator independently rates all nodes and edges on a scale of 1-10, and records disputed cases). Secondly, for entries with a scoring difference of more than 2 points, the team conducts group discussions, referring to the knowledge base (WikiData) to reach a consensus. Finally, the average score of the three annotators is taken as the final baseline data. The Fleiss’ Kappa coefficient for the annotation results is calculated to be 0.85,

\subsection{Tranditional Evaluation}

To evaluate the superiority of \methode compared to traditional methods, this paper designs multiple baselines based on traditional statistical, rule-based, and NLP methods for the evaluation metrics. Below are the descriptions of these baselines:  
\begin{itemize}  
    \item \textbf{Single Concept Accuracy (SCA)  Baseline}:  
        \begin{itemize}  
            \item \textbf{Detect-Fuzzy (Fuzzy Word Detection)}: This method builds a regular expression dictionary for fuzzy qualifiers, and calculates the proportion of fuzzy words in the node name:  
            \begin{equation}  
            \text{FR} = 1 - \frac{\text{Count}_{\text{fuzzy}}(v)}{\text{Count}_{\text{total}}(v)}  
            \end{equation}  
            Where \( \text{Count}_{\text{fuzzy}}(v) = \sum_{i=1}^{n} \mathbf{I}(w_i \in v) \), \( \mathbf{I}(w_i \in v) \) is an indicator function that is 1 if the fuzzy word \( w_i \) appears in the node name \( v \), otherwise 0; \( \text{Count}_{\text{total}}(v) \) is the total number of words in node \( v \) (including fuzzy words); the larger the $\text{FR}$, the fewer fuzzy words, indicating a higher SCA.  

            \item \textbf{Detect-Semantic (Semantic Ambiguity Detection)}: This method calculates the semantic similarity (cosine similarity) between the node name and the most similar word in the domain lexicon. If the similarity exceeds a threshold, it is marked as semantically ambiguous:  
            \begin{equation}  
            \text{SA} = \frac{\sum_{v \in V} \text{SA}(v)}{|V|},  
            \end{equation}
             \begin{equation}
                \text{SA}(v) = \begin{cases}  
                1, & \\\text{if} \ \max_{w \in W} \ \text{Sim}(v, w) > T \\
                0, &
                \\\text{if} \ \max_{w \in W} \ \text{Sim}(v, w) \leq T  
                \end{cases}  
            \end{equation}  
            Where \( v \) is the node name, \( W \) is the set of all words in the domain lexicon, \( \text{Sim}(v, w) \) is the cosine similarity between node \( v \) and word \( w \), \( T = 0.7 \) is the set threshold for similarity. If it exceeds this value, it is considered semantically unclear. The larger the $\text{SA}$, the closer the knowledge base is to the domain lexicon, indicating a higher SCA.  

        \end{itemize}  

    \item \textbf{Hierarchy Relationship Rationality (HRR) Baseline}:  
        \begin{itemize}  
            \item \textbf{Detect-Reverse (Reverse Edge Detection)}: If there is an edge \( A \rightarrow B \) and \( B \rightarrow A \), and the relationships are the same, it is marked as a logical error:  
            \begin{equation}  
                \text{RE} = 1 - \frac{E_{\text{reverse}}}{E_{\text{total}}}  
            \end{equation}  
            Where \( E_{\text{reverse}} \) is the number of reverse edges, and \( E_{\text{total}} \) is the total number of edges in the graph; the larger the $\text{RE}$, the fewer reverse edges, indicating higher HRR.  

            \item \textbf{Detect-Cycle (Cycle Dependency Detection)}: The Tarjan algorithm is used to identify strongly connected components, and the proportion of cyclical paths is calculated:  
            \begin{equation}  
                 \text{CT} = 1 - \frac{E_{\text{cycle}}}{E_{\text{total}}}  
            \end{equation}  
            Where \( E_{\text{cycle}} \) is the number of edges involved in the cycle, and \( E_{\text{total}} \) is the total number of edges in the graph; the larger the $\text{CT}$, the lower the proportion of cyclical paths, indicating weaker logical cycle dependency and higher HRR.  

            \item \textbf{Detect-Anomaly (Statistical Anomaly Detection)}: The frequency distribution of edges is calculated, and if a certain type of relationship (such as "belongs to") deviates significantly from the domain benchmark (Chi-square test), it is deemed an anomaly:  
            \begin{equation}  
                \text{AS} = \sum_{r} \frac{|f_r - f_r^{\text{baseline}}|^2}{f_r^{\text{baseline}}}  
            \end{equation}  
            Where \( f_r \) is the observed frequency of relationship \( r \), and \( f_r^{\text{baseline}} \) is the expected frequency of \( r \) under the domain benchmark. The formula sums the deviations between observed and expected frequencies for all relationships \( r \); if the observed frequency of a relationship deviates significantly from the domain benchmark, it indicates an anomaly in the graph, leading to lower HRR.  
        \end{itemize}  

    \item \textbf{Hierarchy Relationship Exclusivity (HRE) Baseline}:  
        \begin{itemize}  
            \item \textbf{Detect-Cluster (Semantic Clustering Detection)}: For each parent node \( p \) and its child nodes \( \{c_i\} \), the GloVe pre-trained model is used to extract the word vectors of the child node names. Then, hierarchical clustering is performed using the WARD algorithm, and the silhouette coefficient \( S_p \) for each parent node \( p \) is computed. Finally, the semantic clustering score \( SC \) is calculated:  
            \begin{equation}  
                \text{SC} = \frac{\sum_{p \in P} [S_p \geq 0.5]}{|P|}  
            \end{equation}  
            Where when \( SC < 0.5 \), it is determined that the child nodes are semantically scattered.  
        \end{itemize}  

    \item \textbf{Hierarchy Relationship Independence(HRI) Baseline}:  
        \begin{itemize}  
            \item Detect-Redundant (Redundant Edge Detection): If node \( v \) has multiple incoming edges \( {p_1 \rightarrow v, p_2 \rightarrow v} \), and \( p_1 \) and \( p_2 \) have an inheritance relationship in the graph, they are marked as redundant edges. The redundant edge detection score \( \text{RED} \) is calculated as follows:  
            \begin{equation}  
                \text{RED} = 1 - \frac{\text{redundant edges}}{\text{total edges}}  
            \end{equation}  

            \item Detect-Modular (Modular Coupling Detection): The Louvain algorithm is used to detect community structure, and the consistency of node \( v \) with the community of its neighbors is calculated. The modular coupling score \( \text{MOD} \) is calculated as follows:  
            \begin{equation}  \begin{aligned}
                \text{MOD} = \\
                -\frac{\sum_{v \in V} \mathrm{I}(\text{c}(v) = \text{mode}(\text{c}(N(v))))}{|V|}  
            \end{aligned}\end{equation}  
            Where the higher the $\text{MOD}$, the weaker the structural coupling and the higher the independence.  
        \end{itemize}  

\end{itemize}  

\begin{table*}[]
\centering
\label{tab:baseline2:1}
\small
\begin{tabular}{@{}p{1.5cm}p{3.4cm}p{1.8cm}p{1.8cm}p{4.5cm}@{}}
\toprule
Metric                                                                      & Dataset                              & Method              & Score     & Explanation                                                   \\ \midrule
{Single Concept Accuracy (SCA)}   & {MAG-CS.small}      & Detect-Fuzzy     & FR=0.981  & The score is close to 1, indicating that there are almost no fuzzy terms in MAG-CS.small, resulting in very high SCA.                \\
                                                                         &                                    & Detect-Semantic  & SA=0.998  & The score is close to 1, indicating that there are almost no terms outside the CS domain in MAG-CS.small, resulting in very high SCA.             \\ \midrule
Hierarchy Relationship Rationality (HRR)      & {Ali-Taxo.LLM.Reverse} & Detect-Reverse   & RE=0.995  & The score is close to 1, indicating that there are almost no reverse relationships in Ali-KB.LLM.Reverse, resulting in very high HRR.         \\
                                                                         &                                    & Detect-Cycle     & CT=1.000   & The score is 1, indicating that there are no cycles in Ali-KB.LLM.Reverse, resulting in very high HRR.                 \\
                                                                         &                                    & Detect-Anomaly   & AS=0.480  & The score is 0.48, which is below the chi-square critical value of 9.488, indicating no significant difference between observed and expected frequencies, resulting in very high HRR. \\ \midrule
Hierarchy Relationship Exclusivity (HRE)             & Ali-Taxo.LLM.Short               & Detect-Cluster   & SC=0.410  & The score is 0.41, indicating that the clustering of child nodes under the parent node is between good and poor, and HRE may be at a medium level.            \\ \midrule
{Hierarchy Relationship Independence (HRI) }    & {Ali-Taxo.LLM.Short}   & Detect-Redundant & RED=1.000    & The score is 1, indicating that there are no redundant nodes, resulting in very high HRI.                                  \\
                                                                         &                                    & Detect-Modular   & MOD=0.120  & The score is 0.12, which is between the worst (-1) and best (1), indicating that HRI is at a medium level.                  \\ \bottomrule
\end{tabular}
\caption{Results of Traditional Baseline Methods for Evaluating Taxonomy.}

\end{table*}
The experimental results of traditional baselines are presented in Table \ref{tab:baseline2:1}. In these experiments, the domain ontology used in Detect-Semantic is the Computer Science Ontology (CSO), which covers various fields in computer science and includes a large number of computer science terms. In Detect-Anomaly, it is assumed that the relationship frequencies are uniformly distributed. Based on the experimental results, the following conclusions can be drawn:

\textbf{LLM methods have an advantage in dynamic semantic understanding.  
}From the evaluation results, the LLM scored 8.09/10 in concept semantic accuracy (SCA). Although this score is relatively low, it is still better than the evaluation results of traditional methods. Manual verification shows that, on the same dataset, the LLM score for SCA is 7.93/10, while traditional methods score nearly 1 (0.998). This indicates that traditional methods have limitations in identifying semantic issues, especially when dealing with polysemy and contextual dependencies in text. Traditional methods rely on static lexicons, neglecting the multiple meanings and context of words. In contrast, LLM dynamically adjusts through contextual reasoning, effectively handling semantic ambiguity, thus improving evaluation accuracy. Therefore, LLM methods have a clear advantage in dynamic semantic understanding.

\textbf{Traditional methods rely on predefined rules and cannot detect unregistered implicit logical contradictions, whereas LLM methods can dynamically correct them through semantic reasoning.  
}When evaluating relation logic (HRR), the LLM scored 0.12, significantly lower than the high scores (0.995, 1.0) in traditional methods. This difference suggests that traditional methods are overly dependent on fixed rules, focusing mainly on explicit logical relationships, such as detecting contradictory and cyclical relations. However, these rules fail to address complex implicit logical contradictions and unregistered complex relationships, limiting the evaluation effectiveness. In contrast, LLM, through semantic reasoning and contextual understanding, can identify and correct these implicit logical issues, resulting in a more accurate assessment of edge relation logic. Through this dynamic correction mechanism, LLM methods can adapt to varying reasoning scenarios, providing more comprehensive and flexible evaluations.

\textbf{Traditional methods are limited by clustering quality and ontology completeness, unable to handle cross-domain mixed categories, whereas LLM methods can flexibly judge through contextual understanding.  
}In the evaluation of Hierarchy Relationship Exclusivity (HRE), the LLM scored 0.00, significantly lower than the 0.41 achieved by traditional methods. This result demonstrates that LLM can flexibly understand and judge the relationships between nodes through context, especially in handling cross-domain mixed categories, showing strong adaptability and accuracy. Traditional methods, with a heavy reliance on ontologies, have their evaluation effectiveness limited by clustering quality and ontology completeness. They often struggle to provide accurate results when dealing with complex cross-domain data. In contrast, LLM, through dynamic analysis of context, can more precisely handle the exclusivity relationships between nodes, particularly when dealing with knowledge intersections and fuzzy boundaries between domains, exhibiting greater flexibility and judgment capabilities.

\textbf{Traditional methods can only detect explicit redundancy and cannot detect implicit coupling at the semantic level, whereas LLM methods can deeply identify it through semantic analysis.  
}In the Hierarchy Relationship Independence (HRI) evaluation, the LLM scored 6.07, compared to 1 (Detect-Redundant) in traditional methods, indicating a significant gap. Traditional methods mainly rely on detecting explicit redundancy and cannot delve into the semantic level, ignoring implicit coupling or dependency relationships at the semantic level. For example, traditional methods might only identify duplicate nodes in the data but fail to detect potential semantic redundancies or implicit dependencies. LLM, through deep semantic analysis, can identify these implicit coupling relationships from a higher-level structural perspective, enhancing the depth of redundancy detection and expanding the evaluation dimension of structural independence, highlighting its advantages in processing complex data.

\end{document}